\documentclass[letterpaper, 10 pt, conference]{ieeeconf}  
\usepackage{amsmath,amssymb,amsfonts}
\usepackage{algorithmic}
\usepackage{graphicx}
\usepackage{textcomp}
\usepackage{xcolor}
\usepackage{booktabs}
\usepackage{multirow}
\usepackage{tabularx}
\usepackage{hyperref}
\usepackage[linesnumbered,ruled,vlined]{algorithm2e}

\IEEEoverridecommandlockouts
\def\BibTeX{{\rm B\kern-.05em{\sc i\kern-.025em b}\kern-.08em
    T\kern-.1667em\lower.7ex\hbox{E}\kern-.125emX}}
\begin{document}

\title{CHARM: Considering Human Attributes for Reinforcement Modeling
}

\author{Qidi Fang$^{1,\dagger}$, Hang Yu$^{1,\dagger}$, Shijie Fang$^{1}$, Jindan Huang$^{1}$, Qiuyu Chen$^{1}$, Reuben M. Aronson$^{1}$, 
Elaine S. Short$^{1}$ 
\thanks{\textsuperscript{†}These authors contributed equally to this work.}
\thanks{$^{1}$Tufts University School of Engineering, Computer Science. Medford, Massachusetts, United States of America
        {\tt\small \{qidi.fang, hang.yu625917, shijie.fang, jindan.huang, qiuyu.Chen, reuben.aronson, elaine.short\}@tufts.edu}}%
}

\maketitle

\begin{abstract}
Reinforcement Learning from Human Feedback has recently achieved significant success in various fields, and its performance is highly related to feedback quality. 
While much prior work acknowledged that human teachers' characteristics would affect human feedback patterns, there is little work that has closely investigated the actual effects.  
In this work, we designed an exploratory study investigating how human feedback patterns are associated with human characteristics. 
We conducted a public space study with two long-horizon tasks and 46 participants.
We found that feedback patterns are not only correlated with task statistics, such as rewards, but also correlated with participants' characteristics, especially robot experience and educational background. 
Additionally, we demonstrated that human feedback value can be more accurately predicted with human characteristics compared to only using task statistics. 
All human feedback and characteristics we collected, and codes for our data collection and predicting more accurate human feedback are available at 
\url{https://github.com/AABL-Lab/CHARM}. 
\end{abstract}

\section{Introduction}

Reinforcement Learning from Human Feedback (RLHF) offers a solution for robots to learn and adapt their policies to human preferences.
Despite much previous work in learning or simulating human feedback has mentioned that human feedback patterns differ from person to person and may be affected by factors beyond the task itself \cite{yu2023thumbs, huang2024modeling, PANADERO2022100416, ghosal2023effectmodelinghumanrationality}, most prior work assumes that all human feedback is correct, or directly simulates human feedback from reward values with random noises \cite{krening2018interaction}.  
RLHF algorithms require a substantial amount of high-quality human feedback for learning. 
Oversimplifying human feedback patterns will introduce a lab-to-real-world discrepancy and result in a performance drop for the learning algorithms. 
Our key insight is that
\textit{human feedback is not only associated with the task and the agent's performance, but also affected by human characteristics.} 
In this work, we therefore concern ourselves with the problem of investigating the relation between human feedback patterns and human characteristics. 


\begin{figure}[h]
    \centering
    \includegraphics[width=\columnwidth]{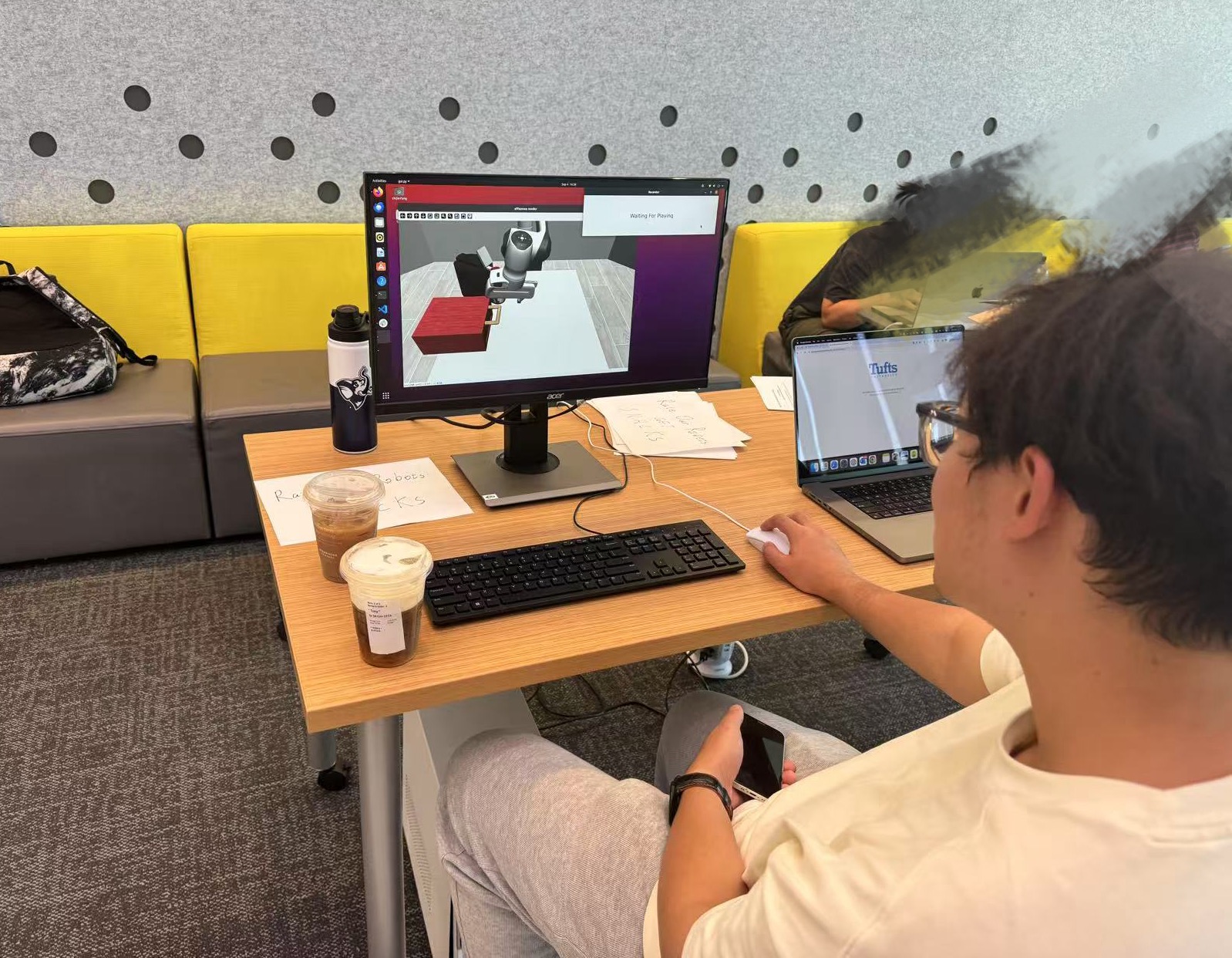}
    \caption{Our public study setup. The experiment was conducted in the atrium of a university building, where we recruited participants through our study setup and snack rewards. A total of 46 participants were recruited. Participants first filled out a questionnaire that covered six domains of human characteristics, and then provided human feedback to two robot tasks.  
    }
    \label{fig:first_page_figure}
\end{figure}

%

RLHF has gained significant attention in recent advancements in robot learning and large model post-training. 
However, RLHF methods are sensitive to data quality and changes in human feedback patterns \cite{10161105}. 
It is widely understood that people's reactions and responses to various situations are influenced by many factors, such as personality \cite{bolmont2001relationships} and rationality level \cite{ghosal2023effectmodelinghumanrationality}. Human factors like teaching style, teaching experience, educational background, robot experience, and trust in robots,  though not definitively linked, could play a role in shaping how individuals provide feedback in the context of teaching robots \cite{PANADERO2022100416}.
Real human feedback is often biased \cite{ibarz2018reward}, erroneous \cite{brawer2023interactive}, delayed \cite{knox2009interactively}, and sometimes even missed \cite{dromnelle2020coping}. 
Different individuals may also give varying feedback for the same observation \cite{kupcsik2018learning}. 
Furthermore, RLHF often requires a large amount of human feedback, which can be costly to collect.
One solution for mitigating the data shortage and reducing the cost is to use simulated human feedback generated by oracles. 
Commonly used oracles, often based on optimal policies \cite{sahni2016policy}\cite{swamy2024minimaximalist} or heuristic functions \cite{griffith2013policy}\cite{du2019provably}\cite{sheidlower2022keeping}, did not account for human factors. 

Despite the recognized importance and broad existence of these dynamic human feedback patterns, few studies have explicitly considered how factors beyond the task affect human feedback \cite{huang2024modeling}.
Assuming users would consistently provide feedback only based on the task or relying on oracles to provide instant, bias-free feedback oversimplifies how humans interact with robots \cite{krening2018interaction}.
These oversimplifications can result in biased outcomes when transitioning from in-lab developments to real world applications.
To bridge this gap, our work focuses on investigating the impact of human characteristics on feedback patterns and exploring if incorporating these attributes can improve predicting human feedback. 


We investigate how human characteristics are associated with human feedback patterns and propose that human feedback values would be more accurately predicted if human characteristics are taken into account. 
We designed a questionnaire capturing six key aspects of human characteristics that are mentioned to be correlated with human feedback patterns in prior work. We conducted a public space study with 46 participants, with each participant providing feedback on a nut assembly task and a coffee preparation task. Our results indicate that while our selected human characteristics have a low correlation with feedback delay, they are fairly correlated with feedback values. 
We found that educational background and robot experience are the most correlated domains while teaching experience and trust in robots have lower correlations. 
In addition, we proposed a novel method for training a human feedback predictor CHARM, Considering Human Attributes for Reinforcement Modeling.
We found that models with human characteristics and task statistics (CHARM) achieved significantly higher accuracy in predicting feedback values compared to models with only task statistics.  
We published all human feedback and human characteristic questionnaire responses, and our code
online at \url{https://github.com/AABL-Lab/CHARM}.

In this work, we demonstrated that our selected human characteristics exert a limited correlation with feedback delay but a pronounced correlation with feedback values. 
While it is well-established that human feedback patterns can vary among individuals \cite{yu2023thumbs, christiano2017deep}, to the best of our knowledge, this is the first work that 
systematically investigated how human characteristics are associated with human feedback and leveraged human characteristics to predict human teaching signals. 
We believe that this work has the potential to reduce
the gap between robot learning in the lab and robot learning
in the wild and benefit the development of RLHF methods.

\section{Background}
\noindent\textbf{Reinforcement Learning from Human Feedback.}
Reinforcement Learning from Human Feedback (RLHF) is a technique that enables agents to learn and adapt their behavior based on feedback provided by humans, rather than relying solely on predefined reward functions \cite{sutton2018reinforcement, yu2024much}.
RLHF has been applied successfully in various fields, particularly in robot learning, where it allows robots to adjust their strategies based on human preferences \cite{liu2022task} \cite{griffith2013policy} \cite{TAN2020113043}, critiques \cite{cederborg2015policy}, and ratings \cite{arumugam2019deep}, and natural language \cite{kuhlmann2004guiding}. Research indicates that human feedback can significantly reduce unnecessary exploration during the early training stages \cite{cruz2015interactive}, leading to more efficient convergence \cite{knox2009interactively}.

\noindent\textbf{Human Oracles in RLHF.}
One major challenge of RLHF is the large volume of human feedback required \cite{corrado2024guided, yu2021active}. Using oracles can significantly reduce the need for human feedback, lowering costs associated with participant recruitment and briefings\cite{wang2020teaching} \cite{reddy2018shared}.
One common approach is to generate oracles from heuristic functions. In this method, a hand-engineered function maps state-action pairs to positive or negative critiques instead of relying on binary feedback directly from users \cite{martins2013heuristically} \cite{rosenfeld2018leveraging} \cite{mehta2024unified}. Another approach generates oracles from optimal policies, where an optimal policy is usually a fully trained model and feedback is given based on the alignment between the current robot policy and the optimal policy \cite{sheidlower2022keeping} \cite{cederborg2015policy}. 
However, despite their convenience, oracles can deviate from real human feedback  \cite{lindner2022humans}. 
Human feedback can be biased \cite{thomaz2006reinforcement}, unreliable\cite{faulkner2020interactive}, and inconsistent \cite{bignold2021evaluation, yu2023thumbs}. Furthermore, human feedback is rarely instantaneous; delays can occur due to distraction \cite{kessler2019active}, or hesitation \cite{loftin2014strategy}. 
Relying on perfect oracles that provide instant, error-free feedback oversimplifies the complexity of human-robot interactions \cite{krening2018interaction}. 
Moreover, human feedback patterns can vary across individuals and often cannot be captured by perfect oracles, which creates a gap between in-lab performance and real-world performance.

\noindent\textbf{Human Characteristics and Human Feedback Patterns.}
Personality plays a role in shaping how people interact with and respond to different situations \cite{bolmont2001relationships}\cite{1513803}. 
The widely used Big Five personality model—comprising openness, conscientiousness, extraversion, agreeableness, and neuroticism—captures these traits \cite{soldz1999big}. 
Individuals with higher conscientiousness are efficient and responsible \cite{mccrae1992introduction} and may provide less delayed feedback. People high in neuroticism may deliver high-quality feedback \cite{pelgrim2014factors}, especially during intensive robotic tasks. In addition to personality, teaching experience and teaching style may also indirectly affect human teaching behavior.
People with formal teaching experience may provide more structured and constructive feedback \cite{georgountzou2019peer} \cite{brinko1993practice}.
Educational background, robot experience, and trust in robots further shape feedback patterns. Individuals with more robot experience or with relevant technical backgrounds may better understand the nuances of interaction with robots \cite{8525515}, providing more precise feedback. Trust in robots could also play a role—those with a higher trust may be more forgiving of robot errors \cite{karli2023if}, resulting in biased feedback.


Our work differs from prior work by proposing factors beyond the task itself will affect human feedback patterns and conducting in a non-lab setting. 
By recruiting people with diverse backgrounds, we aim to investigate how their personal characteristics associate with their feedback patterns and whether we can better predict their feedback by considering their personal characteristics.

\section{Methodology}

Providing feedback is a complex process involving multiple cognitive functions \cite{maior2018workload}.
We argue that human feedback does not solely rely on tasks or robot behaviors, it also depends on the human teachers themselves. 
To explore this, we developed a questionnaire to capture six key domains of human characteristics and conducted a public study to collect both human feedback and their characteristics. 
Moreover, 
we proposed a novel method for training human feedback simulator CHARM, Considering Human Attributes for Reinforcement Modeling.

\subsection{Human Feedback pattern}



Real human feedback differs from simulated human feedback in two key aspects: subjective evaluation values and the timing of feedback delivery. Scalar feedback offers a rich representation by capturing both positive and negative magnitudes while incorporating delay accounts for individual differences rather than assuming a uniform delay distribution \cite{knox2009interactively}. Thus, in this work, we represent human feedback patterns using scalar feedback and delays.
\subsubsection{Delay}
We represent feedback delay as a non-negative number $D$ that denotes the time difference between the end of the robot trajectory and the moment feedback is provided.


\subsubsection{Scalar Value}
The scalar value represents the degree of satisfaction or dissatisfaction in feedback. We used scalar feedback in a five-point likert scale \( V \in \{-2, -1, 0, 1, 2\} \), representing from "Very Dissatisfied" to "Very Satisfied".




\begin{figure}[tb]
    \centering
    \includegraphics[width=0.48\textwidth]{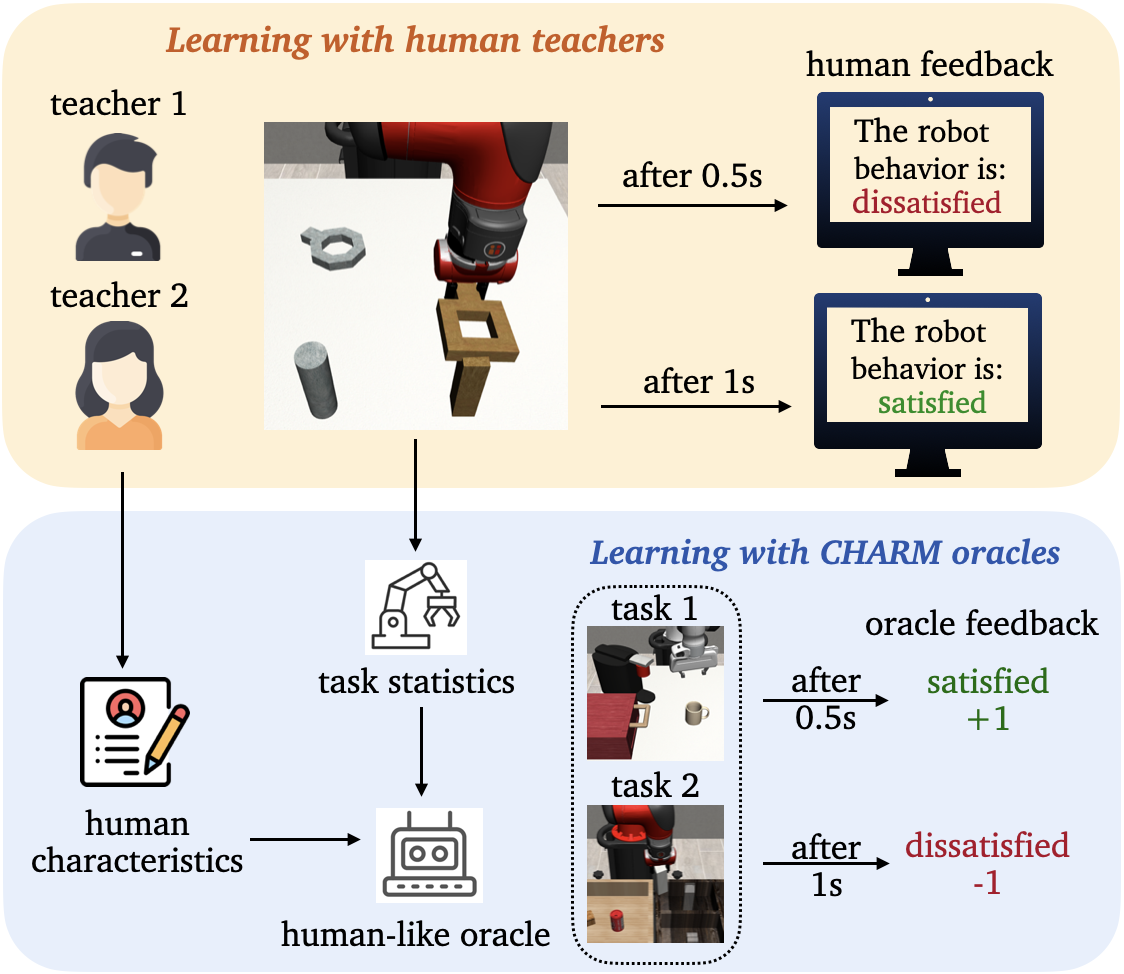}
    \caption{Learning from human teachers vs. learning from CHARM. Human teachers provide feedback with varying delays and responses, while CHARM oracles predict this feedback by incorporating human characteristics and task statistics. 
    }
    \label{fig:oracle_overview}
\end{figure}
\subsection{Questionnaire}
As mentioned in prior work \cite{bolmont2001relationships, 1513803, soldz1999big, mccrae1992introduction, pelgrim2014factors}, human feedback patterns can vary from person to person. 
In this work, we designed a questionnaire to capture the differences in human characteristics among individuals. 
The questionnaire consists of six domains, including trust in robots, robot experience, educational background, teaching experience, teaching style, and personality traits.
The details of the questionnaire we used are as follows:
\begin{itemize}
    


    
    
    \item \textbf{Trust in Robots}: We combined two questions from \cite{charalambous2016development} and one from \cite{nomura2004psychology} to measure participants' trust in robots, focusing on perceived reliability, safety, and concerns about overreliance.
    \item \textbf{Robot Experience}: We adapted two questions from \cite{schaefer2013trust} to evaluate participants' familiarity with robots, emphasizing their prior experiences with both general robots and robotic arms.
    \item \textbf{Educational Background}: We designed three questions to capture participants' academic backgrounds, considering both their highest degree and their proximity to fields such as computer science or robotics.
    \item \textbf{Teaching Experience}: We crafted two custom questions to document participants' formal teaching experience, including roles (e.g., TA or Professor) and the number of years they have taught.
     \item \textbf{Personality Traits}: We employed a 10-item short version of the Big Five Inventory \cite{rammstedt2007measuring} to assess individual personality differences, covering openness, conscientiousness, extraversion, agreeableness, and neuroticism.
    \item \textbf{Teaching Style}: We adapted eight questions from \cite{ertesvaag2011measuring} to capture participants' teaching styles in classroom scenarios—drawing on their actual teaching experiences when available or assuming their approach otherwise—with a focus on relationship building, routine setting, care, praise, and behavior monitoring.

\end{itemize}

While some demographic characteristics, such as age, gender, or cultural background, are very likely to influence feedback patterns; we excluded these domains from our considerations to ensure that our results would not contribute to any form of discrimination.

\subsection{Considering Human Attributes for Reinforcement Modeling}

To further investigate the non-linear relationship between human characteristics and better simulating realistic human feedback, we propose a novel human feedback simulation method CHARM, shown in \autoref{fig:oracle_overview}, where we include human characteristics into training data while simulating human feedback. 



In this work, we use rewards from a reward function as our task statistics. 
For a given state $s$, action $a$, and a reward function $\mathbb{R}$, the reward of the state-action pair is:
$\mathbb{R}(s,a) \rightarrow r$.
We fixed the length of a robot trajectory to 30 state-action pairs.
The accumulative reward over a trajectory $\sigma$ can be calculated as
\( \hat{r} = \sum_{k=1}^{k=30} r_k \).
The feedback oracle $\mathcal{F}$ takes two inputs: Human Characteristics $\overrightarrow{hc}$ and task statistics $\hat{r}$, and outputs the predicted human feedback $f^*$, described as $
\mathcal{F}(\overrightarrow{hc}, \hat{r}) \rightarrow f^*
$.

\begin{algorithm}[htbp]
\caption{Training Protocol}\label{alg:training_protocol}
Pre-train agents for feedback collection\\
Obtain task statistics (we used a reward function $\mathbb{R}$)\\ 
\For {N participants}{
Vectorizing characteristic questionnaire responses $\overrightarrow{hc}$\\
    \While {robot task not ends}{
        Sample a trajectory $\tau$ from an agent\\
        Calculate task statistic $\hat{r} = \mathbb{R}(\tau)$\\
        Ask participant to provide feedback $f$\\
        Predict human feedback $\mathcal{F}(\overrightarrow{hc}, \hat{r}) \rightarrow f^*$ \\
        Train the feedback prediction model $\mathcal{F}$ by minimizing  
        $\mathcal{L} = \mathcal{L}_{\text{cls}} +  \mathcal{L}_{\text{reg}}$
    }
}
\end{algorithm}

We choose to use a Multilayer Perceptron (MLP) to learn the oracle \( \mathcal{F} \). The training protocol is detailed in \autoref{alg:training_protocol}. We assume that for similar task statistics, users will provide similar feedback values and that the distribution of delay remains stable for the same user. Consequently, the oracle's training objective is to minimize the discrepancy between the predicted and the ground-truth human feedback patterns in terms of both delay and value.
We use two loss components: a classification loss for the feedback value and a regression loss for the delay. Let
$\mathcal{L}_{\text{cls}} = \text{CrossEntropy}(Value, Value^{\text{gt}})$
be the loss for the classification task, and
$\mathcal{L}_{\text{reg}} = \text{MSE}(Delay, Delay^{\text{gt}})$
be the loss for the regression task. The overall loss function is then given by:
$\mathcal{L} = \mathcal{L}_{\text{cls}} +  \mathcal{L}_{\text{reg}}$.






\section{Experimental Setup}

We conducted a public space study with 46 participants in total.
This allowed us to recruit participants with a broad range of characteristics. 

\subsection{Task Setup}
\label{sec:ts}

\begin{figure}[tbp]
    \centering
    \includegraphics[width=\columnwidth]{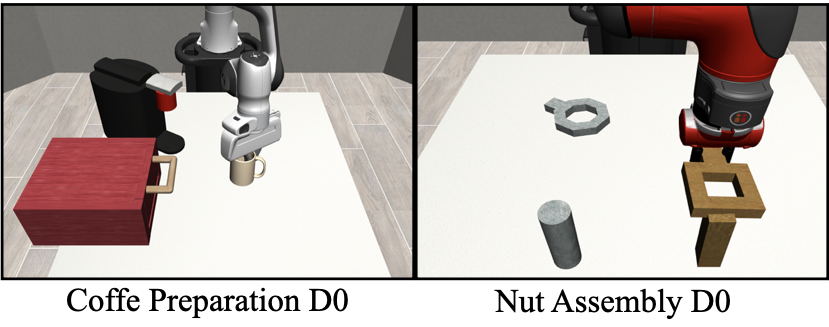}
    \caption{Robotic tasks from mimicgen \cite{mandlekar2023mimicgen} in robosuite simulator\cite{zhu2020robosuite}.}
    \label{fig:robot_tasks}
\end{figure}

Our experiment used two tasks from mimicgen \cite{mandlekar2023mimicgen}, nut assembly, and coffee preparation, as shown in \autoref{fig:robot_tasks}. 
For each task, we pre-trained agents with different steps using the BC-RNN algorithm \cite{mandlekar2021matters}, leveraging 1,000 human expert demonstrations from \cite{mandlekar2023mimicgen}.
For each task, we trained three agents, 500 epochs (minimally trained), 1,000 epochs (medium-trained), and 2,000 epochs (well-trained), to generate trajectories with diverse proficiencies. During the experiment, participants were shown six randomly sampled trajectories (one trajectory per agent, three agents for two tasks). 
Additionally, we used the Density-Based Reward Modeling algorithm \cite{gleave2022imitation} to extract the reward function from expert demonstrations, which provided rewards and served as the task statistics for each task.
\subsection{GUI for Standardized Feedback Collection}

We developed a graphical user interface (GUI) to standardize and streamline data collection. The GUI consists of two components: the "state window" and the "noticing window". During robot execution, the state window remained continuously displayed. Upon completing a task, the noticing window appeared with a blinking prompt, instructing the participant to provide a rating. This window displayed five emoji buttons labeled from "Very Dissatisfied" to "Very Satisfied," corresponding to a five-point scale in a range of $[-2, 2]$. Additionally, the system recorded the response latency as the elapsed time between the onset of the noticing window and the participant's rating submission, thereby capturing feedback delay. Once a rating was submitted, the noticing window disappeared, and the state window was reactivated for the subsequent trajectory execution. If no response was provided, the noticing window remained visible for a maximum of 5 seconds before automatically proceeding to the next trajectory.

\subsection{Procedure}

The experiment was conducted in an open public space with 46 participants. First, participants filled out a consent form and then completed the human characteristics questionnaire. Before providing feedback, we provided participants with an overview of the tasks that the robot needed to complete. Participants were instructed to rate the robot entirely based on their judgment.
During the experiment, the tasks were rendered on-screen while participants provided feedback via the previously described GUI. For both the nut assembly and coffee preparation tasks, each participant rated three trajectories. For each trajectory—comprising approximately 500 to 800 steps—the system recorded task statistics, feedback values, and feedback delays. The robot paused every 30 actions to prompt for feedback via the GUI; if no response was given within 5 seconds, the system automatically proceeded to the next set of actions.

\begin{figure*}[htbp]
    \centering
    \includegraphics[width=\textwidth]{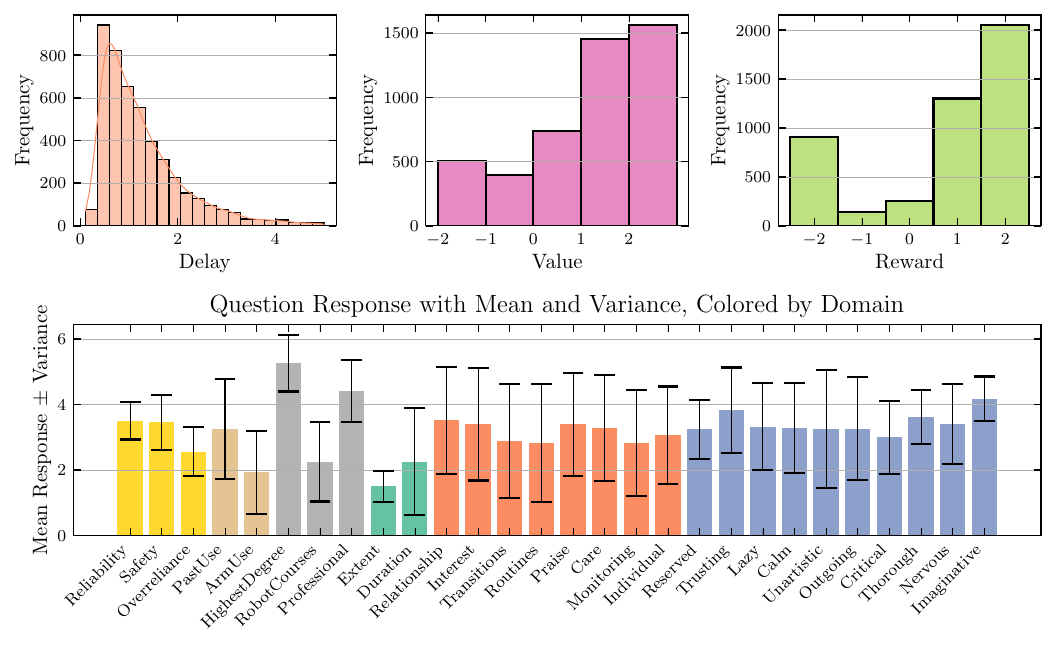}
    \caption{
    Overview of distributions of feedback delay, feedback value, rewards, and human characteristic questionnaire responses for all participants.  
    Top Row (left to right): 
    (1a) Feedback Delay distribution;
    (1b) Feedback Value distribution;
    (1c) Binned reward distribution.
    Bottom Row (2a) is the mean and standard deviation of the responses across all the questions in the human characteristic questionnaire. Each bar corresponds to one question (e.g., Q1 → “Reliability”), and each x-axis label is a word that summarizes the question. Questions from the same human characteristics domain share a common color. From left to right: Trust in Robot, Robot Experience, Education Background, Teaching Experience, Teaching Style, and Personality.
    }
    \label{fig:distribution}
\end{figure*}

\section{Results}
\label{sec:results}

In this section, we first present the data we collected from 46 participants in our public space study.
Our goal was to investigate the correlation between human characteristics and human feedback patterns. 
We found that robot experience and educational background are fairly correlated with feedback accuracy and feedback values. 
We proposed CHARM, a human feedback generation method that considers human characteristics.
CHARM  achieved a higher accuracy in predicting human feedback values compared to the method that only considers task statistics.      


\subsection{Collected Human Feedback and Questionnaire Responses}
\label{sec:distribution}
We collected data from 46 participants, consisting of human characteristics questionnaire responses, task statistics, and feedback delay and value. 
We show our collected data in \autoref{fig:distribution}. 

As shown in the top row, participants’ delays are tightly clustered around one second, which supports the conclusion in prior work \cite{hockley1984analysis}.
The reward distribution broadly aligns with the human feedback distribution. 
We computed feedback accuracy for each participant by considering \(\{-2, -1\}\) as negative, \(\{1, 2\}\) as positive, and \(\{0\}\) as neutral.
On average, participants achieved an accuracy of approximately \(51.79\%\) with a standard deviation of \(10.90\%\), and the highest individual accuracy reached \(74.50\%\). 
We computed the Pearson correlation coefficient \cite{schober2018correlation} between the reward and human feedback values and delays, respectively, yielding
\[
\rho_{r,\text{value}} \approx 0.563, p < 0.001;
\quad
\rho_{r,\text{delay}} \approx -0.004, p = 0.769.
\]

These results indicate a strong linear relationship between reward and feedback value and virtually no correlation between reward and delay. 
We adapted the correlation of 0.563 as a reference point when evaluating the correlation between human characteristics and feedback patterns.
The bottom row of Figure~\ref{fig:distribution} presents the responses of the questionnaire. From left to right, one color represents one domain of our questionnaire, and the x-axis shows the words that briefly summarize questions. 
The mean and standard deviation for each question vary largely, reflecting the diverse backgrounds among participants.

\label{sec:descriptive_analysis}

\begin{figure*}[htbp]
    \centering
    \includegraphics[width=0.95\textwidth]{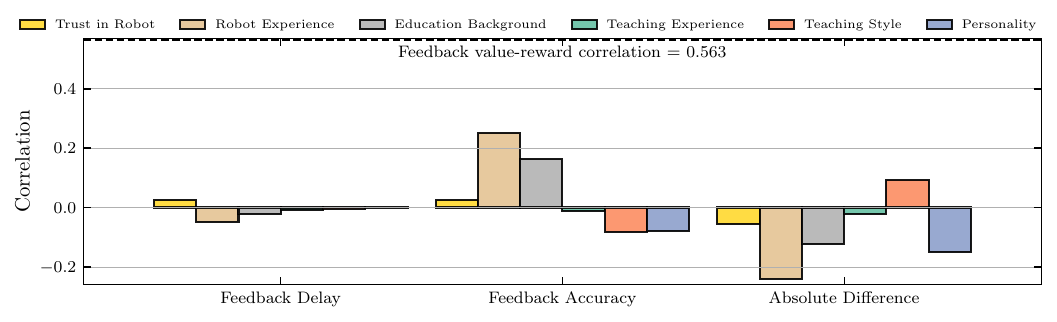}
    \vspace{-5pt}
    \caption{
    Domain-wise correlations between human characteristics and feedback delay, feedback accuracy, and absolute difference 
    Each bar represents one human characteristic domain.
    Given that the feedback value-reward correlation is 0.563 (as the dotted line), correlations exceeding 0.1 are considered noteworthy.
    The figure indicates that there is no substantial linear correlation between any human characteristic domain and feedback delay. 
    Robot experience and educational background have the most noticeable correlations with feedback accuracy and absolute difference.
    }
    
    \label{fig:correlation}
\end{figure*}
\subsection{Correlations between Human Feedback Pattern and Human Characteristics}
\label{sec:correlation_analysis}

In this subsection, we analyzed the correlation between human characteristics and feedback patterns—namely, feedback values and delays. 
We used the correlation between feedback values and rewards (0.563) as the ceiling.   
Since human feedback has been broadly used in RLHF as reward signals and has achieved many successes, 
the correlation between rewards and human feedback is a reasonable upper boundary to assess the significance of the correlation between human characteristics and human feedback.

We calculated the Pearson correlation between each domain of human characteristics and feedback delay, feedback accuracy, and absolute difference. 
Feedback accuracy was calculated as described in the previous subsection, representing the quality of human feedback.
The \textit{absolute difference} is the absolute difference between feedback value and reward value, measuring the degree to which feedback deviates from the ground truth.
We show the results in Figure~\ref{fig:correlation}.
Delay-wise, we did not find any significant correlation between any domain of the questionnaire and feedback delay. 
We found some domains fairly correlated with feedback accuracy and absolute difference.
Robot experience and educational background exhibit correlation coefficients of 0.25 and 0.16 with feedback accuracy. 
Compared to the reward-value correlation (0.563), a coefficient above 0.1 is fairly notable. This suggests that participants with relevant educational backgrounds or greater robot experience tend to provide feedback that is more accurate in direction (e.g., positive vs. negative).
Furthermore, robot experience and educational background show negative correlations of -0.24 and -0.12, respectively, with the absolute difference. This indicates that participants with more robot experience or a relevant educational background also tend to provide feedback that is closer in value to the ground truth, reflecting a higher degree of numerical precision.

Our correlation analysis indicates that our selected human characteristic domains have a fair correlation with feedback values but little correlation with feedback delay. 
In our selected domains, robot experience and educational background are the most relevant.

\subsection{Including Human Characteristics Improves Feedback Prediction Accuracy}
\label{sec:simulated_feedback_provider_evaluation}
We demonstrate that integrating human characteristics would be beneficial for training human feedback oracle by showing CHARM outperforms the baseline oracle that trained with only task statistics as input. 
In the previous subsection, we did not find any noticeable correlation between human characteristics and human feedback delay.
Thus,
in this subsection, we will focus more on feedback values. 
We first introduce how we trained CHARM and the baseline, and then introduced our results. 

\noindent\textbf{CHARM Training Setup}
As noted in Section~\ref{sec:distribution}, our dataset shows an unbalanced distribution in feedback values. 
To improve the training quality, we applied 
a Box-Cox transformation \cite{SIBONO2025343704} to the delay data and applied
a weighted loss function during training to assign higher weights to minority classes in the feedback value distribution. 
We performed k-fold cross-validation (in this work, k=10) by randomly splitting our \(4655 \) data points into 90\% training and 10\% validation sets, with a new random seed every time.
We demonstrated the models' performance on two feedback value scales. 
One is the original value scale, a five-point scale, ranging from -2 to 2. The other one is a widely used feedback value scale, binary feedback, with only positive and negative.  
We adapted the data set for training the models to predict binary feedback by classifying \(-2\) and \(-1\) as negative, classifying \(1\) and \(2\) as positive, and dropping the 0s. 
We had three methods for predicting feedback values.
Random prediction: the model randomly selected a feedback value (expected accuracy is 20\% and 50\%).
Task statistics only: the model was trained to predict feedback values using corresponding task statistics. 
Human characteristics and task statistics (CHARM): the model was trained to predict feedback values using both corresponding task statistics and human characteristics.  



\begin{figure}[htbp]
    \centering
    \includegraphics[width=0.95\columnwidth]{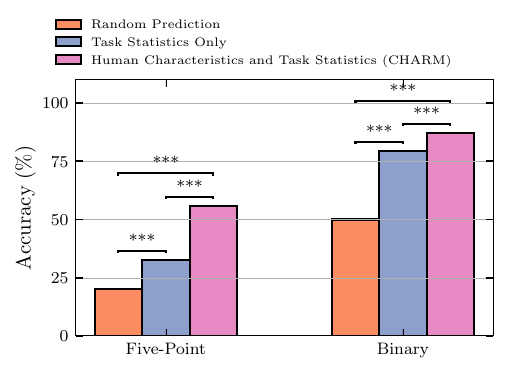}
    \caption{Comparison of feedback prediction accuracy. 
    Left: Feedback prediction on a five-point scale (ranging from -2 to 2).
    Right: Binary feedback prediction (positive vs.\ negative).
    Each bar represents the average accuracy across a 10-fold cross-validation. 
    Results indicate that models trained with human characteristics and task statistics combined (CHARM) significantly outperform models trained with task statistics only ($p < 0.001$, for all paired $t$-tests).}
    \label{fig:comparison_accuracy}
\end{figure}

\noindent\textbf{Results}
We first examined feedback delay simulation. 
The model trained solely on task statistics (reward) yielded a Mean Squared Error (MSE) of \textit{0.30} and an $R^2$ of \textit{0.19} to the true feedback delay, while the model trained with task statistics and human characteristics has an MSE of \textit{0.37} and an $R^2$ of \textit{0.02}.
Both models performed poorly and including human characteristics showed minimal improvement. These results are consistent with our earlier correlation analysis.
We show our feedback value results on two feedback value scales in Figure~\ref{fig:comparison_accuracy}. 
Under the original five-point scale setting, 
the model trained with only task statistics has an accuracy of 32.59\% and the model trained with task statistics and human characteristics (CHARM) has an accuracy of 55.83\%. 
Both models significantly outperformed the random prediction, and 
including human characteristics significantly improved the accuracy of predicting five-point scale feedback ($t = 50.42$, $p < 0.001$).
Under the binary feedback setting, 
the model trained with only task statistics has an accuracy of 79.35\% and CHARM has an accuracy of  87.03\%. 
The improvement of including human characteristics in predicting binary feedback was statistically significant ($t = 27.80$, $p < 0.001$).
Overall, our results suggest that our selected human characteristics do associate with feedback values, and including these characteristics could help the machine learning model better capture the underlining human feedback value patterns. 



\section{Discussion}
We found that there is a fair correlation between the selected human characteristics and human feedback values, especially on robot experience and educational background, while human feedback delay is not correlated with our selected human characteristics.
We also found that human feedback values can be more accurately predicted if we include human characteristics into training. 


One interesting finding is that while there were no strong correlations between the selected human characteristics and human feedback values, our experiment results demonstrated that incorporating human characteristics into training leads to a significant improvement in modeling human feedback patterns. 
We assume this is because of two reasons. 
One is that the relationship between human characteristics and feedback patterns might not be linear, and multiple domains of human characteristics can jointly affect human feedback.
The other one is that human characteristics are only one of the factors that affect human feedback values, and the human feedback values also largely depend on the agent's true performance.  

One limitation of this work is that we paused the robots when asking for feedback to more accurately record feedback delay.
However, the feedback delay data might not be representative for continuous interactions. Future work should further extend this work to collect feedback from both paused and un- paused conditions, improving the representativeness of the data. 

\section{Conclusion}
In this work, we selected six categories of representative human characteristics to investigate the relationship between human characteristics and human feedback patterns. 
We conducted our study in a public space with 46 participants who randomly passed by our setup. 
We found that our selected human characteristics have relatively high correlations with human feedback values, while the correlation with human feedback delay is low. 
We also proposed a novel feedback-simulating method CHARM and demonstrated that a machine learning mode can more accurately predict true human feedback values if human characteristics were included in the training data. 
Our method offers a new perspective for better predicting human feedback and developing more robust RLHF algorithms, highlighting that human teaching depends not only on the task itself but also on the individual differences among human teachers.


\bibliographystyle{IEEEtran}
\bibliography{references}

\vspace{12pt}

\end{document}